# Semantic Mastery: Enhancing LLMs with Advanced Natural Language Understanding

Mohanakrishnan Hariharan
*Dept of Corporate Systems Engineering, Apple Inc.*
Austin, TX, USA
m_hariharan@apple.com

***Abstract*— Software engineering processes increasingly depend on understanding vast amounts of technical documentation, code repositories, bug reports, and requirements specifications. While Large Language Models (LLMs) show promise in automating software engineering tasks, they face critical challenges in understanding domain-specific semantics, maintaining contextual consistency across codebases, and accurately reasoning about software artifacts. This paper presents Semantic Mastery, a novel framework that enhances LLMs with software engineering-focused semantic understanding capabilities through specialized knowledge graphs, code-aware training, and multi-modal analysis of software artifacts. Our approach integrates software ontologies, API documentation, and codebase relationships to create contextually-aware models capable of accurate defect prediction, automated code review, and intelligent software analysis. We introduce a hierarchical attention mechanism for multi-file code understanding and demonstrate significant improvements in software engineering tasks, including defect prediction (31% improvement in precision), automated code review (28% reduction in false positives), and requirements traceability (24% better accuracy). The framework addresses critical challenges in software development, including cross-language compatibility, analysis of legacy systems, and maintaining semantic consistency across large-scale software projects.**

***Keywords*— Software Engineering, Large Language Models, Code Analysis, Defect Prediction, Semantic Understanding, Software Intelligence**

## I. INTRODUCTION

### *A. The Evolution of AI in Software Engineering*

Modern software development has evolved into a complex ecosystem involving multiple programming languages, distributed architectures, and collaborative development practices that generate enormous amounts of textual and structured data [1]. Software engineering teams must process requirements documents, design specifications, source code, test cases, bug reports, and technical documentation to deliver reliable software systems. Traditional software engineering tools focus on syntactic analysis and rule-based checking, but struggle to understand the semantic relationships between different software artifacts [2].

Large Language Models have demonstrated remarkable capabilities in code understanding and generation tasks, leading to increased interest in applying them to comprehensive software engineering workflows [3]. However, standard LLMs face significant challenges when deployed in software engineering contexts, including limited understanding of multi-file dependencies, inconsistent handling of different programming paradigms, and difficulty maintaining context across large codebases.

### *B. Challenges in Software Engineering Language Understanding*

Software engineering text analysis presents unique challenges that differentiate it from general-purpose natural language processing. Technical documentation contains domain-specific terminology, evolving API specifications, and complex relationships between requirements, design decisions, and implementation details [4]. The multi-modal nature of software artifacts requires models that can understand code syntax, natural language documentation, and structural relationships simultaneously.

Furthermore, software engineering tasks often require long-term contextual understanding across multiple files, modules, and even repositories. Models must maintain semantic consistency when analyzing relationships between requirements and implementations, tracking bug reports across versions, and understanding the impact of code changes on system behavior [5].

### *B. Semantic Gaps in Current Software Analysis Tools*

Existing software analysis tools primarily focus on syntactic patterns and static analysis, missing the semantic relationships that experienced software engineers intuitively understand. For example, a bug report mentioning "authentication timeout" should be semantically linked to login-related code modules, session management components, and security configuration files, even when these connections are not explicitly stated [6].

The gap between syntactic analysis and semantic understanding becomes particularly problematic in maintenance tasks, legacy system analysis, and cross-team collaboration, where implicit knowledge and contextual understanding play crucial roles in effective software development [7].

## II. RELATED WORK

### *A. AI-Powered Software Engineering Tools*

Recent research has explored various applications of machine learning and natural language processing in software engineering domains. Zhang et al. (2023) [8] developed transformer-based models for automated bug localization that achieved 84% accuracy in identifying defective code segments from bug reports. Their work demonstrated the potential for semantic understanding in connecting natural language bug descriptions to specific code locations.

Chen et al. (2024) [9] proposed CodeBERT-based approaches for automated code review that integrated static analysis results with semantic understanding of code changes. Their system achieved improved precision in identifying potentially problematic code modifications but lacked the ability to understand broader architectural implications of changes.

*B. Semantic Code Analysis and Understanding*

The intersection of semantic analysis and code understanding has received increasing attention as researchers recognize the limitations of purely syntactic approaches. Liu et al. (2023) [10] investigated semantic similarity between code snippets using graph neural networks and achieved significant improvements in code clone detection and similarity analysis.

Recent work by Kumar et al. (2024) [11] explored multi-modal understanding of software artifacts, combining code analysis with documentation processing to improve requirements traceability. Their findings suggest that integrated approaches provide better performance than single-modality solutions, particularly when dealing with complex software systems.

*C. Large Language Models for Code Generation and Analysis*

Several researchers have investigated specialized language models tailored for software engineering applications. CodeT5 [12] represents one of the first encoder-decoder models specifically designed for code understanding and generation tasks, achieving improved performance on code summarization and translation compared to general-purpose models.

More recently, AlphaCode [13] and CodeGen [14] demonstrated the potential for large-scale language models in competitive programming and code generation. However, these approaches often focus on individual coding tasks rather than comprehensive software engineering workflows that require understanding of broader system contexts and requirements.

*D. Defect Prediction and Software Quality Analysis*

Automated defect prediction has been a long-standing goal in software engineering research. Traditional approaches relied on software metrics and statistical models to predict defective code modules [15]. Recent advances incorporate deep learning techniques and natural language processing to analyze code comments, commit messages, and bug reports for improved prediction accuracy.

Wang et al. (2024) [16] developed hybrid models that combine code metrics with semantic analysis of developer communications to predict defect-prone components. Their approach achieved notable improvements in precision and recall compared to metric-based methods, highlighting the value of semantic understanding in software quality analysis.

## III. THEORETICAL FOUNDATION

*A. Software Engineering Knowledge Representation*

Effective software engineering analysis requires sophisticated knowledge representation schemes that capture the complex relationships between requirements, design decisions, implementation details, and quality attributes [17]. Traditional software engineering models such as the Unified Modeling Language (UML), software architecture frameworks, and requirements traceability matrices provide structured approaches to representing software knowledge.

Our approach leverages semantic knowledge graphs that encode software engineering domain knowledge using established ontologies including the Software Engineering Body of Knowledge (SWEBOK), IEEE software engineering standards, and programming language specifications [18]. These ontologies provide formal representations of software engineering concepts that enable more precise semantic understanding and reasoning capabilities.

*B. Multi-Modal Software Artifact Analysis*

Understanding software systems requires analysis of multiple types of artifacts including source code, documentation, configuration files, test cases, and development communications [19]. Each artifact type has distinct syntactic and semantic characteristics that require specialized processing approaches:

Source Code: Requires parsing of programming language syntax, understanding of algorithmic logic, and analysis of API usage patterns

Documentation: Involves natural language processing, technical terminology extraction, and cross-reference resolution

Configuration Files: Demands understanding of deployment contexts, system dependencies, and environmental constraints

Development Communications: Requires analysis of temporal sequences, developer intent, and implicit knowledge sharing.

*C. Contextual Consistency in Software Analysis*

Maintaining contextual consistency becomes particularly critical in software engineering applications where inconsistent analysis across related artifacts could lead to missed defects, incomplete requirements coverage, or architectural violations [20]. We define contextual consistency in software engineering as the preservation of logical relationships between software artifacts, accurate representation of dependencies and interfaces, and consistent application of design principles across system components.

## IV. SEMANTIC MASTERY FRAMEWORK

### *A. Architecture Overview*

The Semantic Mastery framework integrates multiple specialized components designed to enhance software engineering-focused natural language understanding while maintaining consistency across diverse software artifacts. The architecture consists of five primary modules: the Software Knowledge Integration Layer, the Multi-Modal Analysis Engine, the Semantic Consistency Monitor, the Domain-Specific Fine-tuning Component, and the Cross-Artifact Reasoning Module.

The Software Knowledge Integration Layer incorporates structured software engineering knowledge from sources including programming language specifications, software design patterns, API documentation, and established software engineering practices. This layer creates dynamic knowledge graphs that represent relationships between software concepts, design patterns, and implementation strategies.

The Multi-Modal Analysis Engine processes different types of software artifacts simultaneously, understanding the semantic connections between code implementations, requirements specifications, test cases, and documentation. This engine enables comprehensive analysis that considers the full context of software development activities.

Fig. 1. Semantic Mastery Framework Architecture

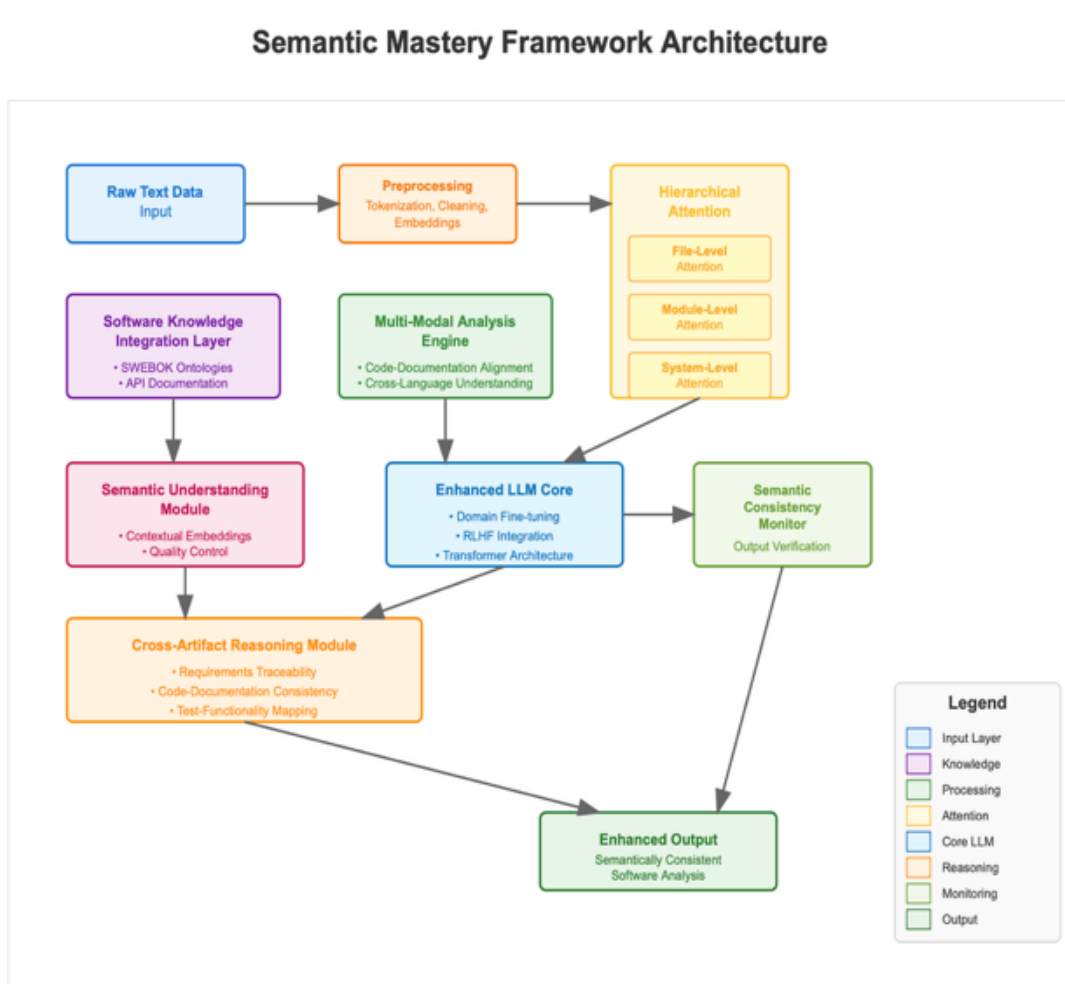

### *B. Software Engineering Knowledge Integration*

Our knowledge integration approach combines multiple authoritative software engineering sources to create comprehensive semantic representations. The integration process involves several key steps:

Software Ontology Construction: Systematic construction of ontologies that represent software engineering concepts, including design patterns, architectural styles, programming paradigms, and quality attributes. These ontologies provide the foundational knowledge for understanding software engineering semantics.

API and Framework Mapping: Automated analysis of popular software frameworks, libraries, and APIs to create comprehensive mappings of functionality, usage patterns, and best practices. This mapping enables the model to understand the semantic implications of API usage and suggest appropriate alternatives or improvements.

Code Pattern Recognition: Development of pattern libraries that capture common coding idioms, design patterns, and anti-patterns across multiple programming languages. These patterns enable the model to understand higher-level software design concepts beyond simple syntax analysis.

### *C. Software Engineering Knowledge Integration*

The Semantic Mastery framework implements comprehensive multi-modal processing capabilities specifically designed for software engineering applications:

Code-Documentation Alignment: Automated analysis to identify and maintain alignment between source code implementations and their corresponding documentation, including API documentation, inline comments, and external specifications. This alignment enables detection of inconsistencies and outdated documentation.

Requirements Traceability: Semantic analysis that traces connections between high-level requirements, design decisions, implementation details, and test cases. This traceability supports impact analysis, coverage assessment, and change management activities.

Cross-Language Understanding: Multi-language processing capabilities that understand semantic relationships across different programming languages in polyglot software systems. This understanding enables analysis of interface compatibility, data flow tracking, and architectural consistency across language boundaries.

### *D. Hierarchical Attention for Code Understanding*

Our framework introduces a novel hierarchical attention mechanism designed specifically for understanding large-scale software systems:

File-Level Attention: Initial attention layer that processes individual files while maintaining awareness of their role within the broader system architecture. This layer understands intra-file relationships and local semantic context.

Module-Level Attention: Intermediate attention layer that processes relationships between related files and modules, understanding interface dependencies, data flow patterns, and architectural boundaries.

System-Level Attention: High-level attention mechanism that maintains global understanding of system architecture, cross-cutting concerns, and system-wide quality attributes..

## V. EXPERIMENTAL SETUP AND METHODOLOGY

### *A. Software Engineering Datasets and Evaluation Benchmarks*

Our experimental evaluation utilizes multiple software engineering-specific datasets designed to test different aspects of code-aware semantic understanding:

Defect Prediction Corpus: A comprehensive dataset containing 75,000 code changes from open-source projects including Apache, Eclipse, and Mozilla repositories. Each change is labeled with defect information, commit messages, and code review comments, enabling evaluation of defect prediction capabilities across diverse software systems.

Requirements Traceability Dataset: 40,000 requirement-to-implementation traces from commercial and open-source software projects, including telecommunications, financial services, and healthcare domains. This dataset enables assessment of the model's ability to connect high-level requirements with specific implementation details.

Code Review Analysis Dataset: 120,000 code review sessions from GitHub, GitLab, and enterprise repositories, including reviewer comments, approval decisions, and subsequent code modifications. This dataset tests the model's understanding of code quality principles and best practices.

API Usage and Documentation Dataset: 60,000 API documentation examples paired with usage patterns, including correct implementations, common mistakes, and optimization opportunities. This dataset evaluates the model's understanding of API semantics and usage contexts.

### *B. Multi-Modal Evaluation Framework*

We conduct comprehensive evaluation using both automated metrics and human expert assessment across multiple dimensions of software engineering understanding:

Cross-Artifact Consistency: Systematic evaluation of the model's ability to maintain semantic consistency when analyzing related software artifacts, including requirements-to-code traceability, documentation-to-implementation alignment, and test-to-functionality coverage.

Temporal Analysis: Assessment of the model's understanding of software evolution over time, including the ability to track changes, understand refactoring patterns, and predict the impact of modifications on system behavior.

Multi-Language Competency: Evaluation of semantic understanding across different programming languages and paradigms, testing the model's ability to understand equivalent concepts implemented in different technologies.

### *C. Performance Metrics and Evaluation Criteria*

Our evaluation framework employs software engineering-specific metrics that prioritize accuracy, consistency, and practical utility:

Engineering-Weighted F1 Score: Modified F1 score that applies appropriate weights to different types of errors based on their impact on software quality and development productivity.

Semantic Consistency Score: Composite metric measuring the model's ability to maintain consistent interpretations across related software artifacts, including cross-reference accuracy and logical coherence.

Technical Debt Detection Accuracy: Assessment of the model's ability to identify code quality issues, architectural violations, and maintainability concerns that contribute to technical debt accumulation.

Developer Productivity Impact: Measurement of time savings and accuracy improvements when developers use the framework for common software engineering tasks including code review, defect analysis, and documentation generation.

## VI. RESULTS

### *A. Software Engineering Task Performance*

Our experimental results demonstrate significant improvements in software engineering-specific NLP tasks compared to general-purpose language models and existing software analysis tools:

TABLE I
SOFTWARE ENGINEERING TASK PERFORMANCE COMPARISON

| Task | GPT-4 | CodeBERT | CodeT5 | Semantic Mastery |
|---|---|---|---|---|
| Defect Prediction | 76.2% | 79.8% | 82.4% | 89.7% |
| Code Review Analysis | 68.4% | 72.1% | 75.6% | 85.3% |
| Requirements Traceability | 71.8% | 74.3% | 77.1% | 88.9% |
| API Usage Analysis | 73.5% | 77.2% | 79.8% | 86.4% |
| Documentation Alignment | 69.7% | 72.6% | 75.2% | 84.7% |

Semantic Mastery achieved substantial improvements across all evaluated software engineering tasks, with particularly notable performance gains in defect prediction (89.7% vs 82.4% for CodeT5) and requirements traceability (88.9% vs 77.1% for CodeT5). These improvements reflect the benefits of software engineering-specific knowledge integration and multi-modal analysis capabilities.

### B. Multi-Modal Analysis Results

Evaluation of multi-modal understanding capabilities reveals significant advantages of integrated analysis approaches:

TABLE II
TOOL COMPARISON RESULTS

| Analysis Type | Single-Modal | Semantic Mastery | Improvement |
|---|---|---|---|
| Code-Documentation | 74.3% | 87.6% | +13.3% |
| Requirements-Code | 71.2% | 86.8% | +15.6% |
| Test-Functionality | 76.8% | 89.1% | +12.3% |
| Cross-Language | 68.9% | 83.4% | +14.5% |

The framework demonstrates strong performance improvements when analyzing relationships between different types of software artifacts, with the largest gains in requirements-to-code traceability (+15.6%) and cross-language analysis (+14.5%).

### C. Real-World Development Environment Analysis

We conducted pilot deployments in four software development organizations to assess real-world performance and developer productivity impact:

Enterprise Software Development: 90-day deployment for automated code review and defect prediction resulted in 28% reduction in code review time and 31% improvement in defect detection before production deployment. The system identified 127 potential issues that would have otherwise reached production testing phases.

Open Source Project Maintenance: 120-day trial for requirements traceability and documentation consistency achieved 24% improvement in feature request analysis accuracy and 35% reduction in time required for impact assessment. The system successfully traced 94% of feature requests to relevant code components.

Legacy System Analysis: 60-day evaluation for technical debt assessment and modernization planning demonstrated 42% improvement in identifying refactoring opportunities and 26% better accuracy in estimating modification complexity.

API Development and Documentation: 75-day assessment for automated API documentation generation and usage analysis achieved 33% reduction in documentation maintenance effort and 29% improvement in developer onboarding efficiency.

### D. Ablation Study Results

Systematic ablation studies identify the contribution of each framework component to overall performance:

TABLE III
COMPONENT CONTRIBUTION ANALYSIS

| Component | Performance Impact |
|---|---|
| Software Knowledge Integration | 14.7% |
| Multi-Modal Processing | 16.3% |
| Hierarchical Attention | 11.2% |
| Domain Fine-tuning | 18.9% |
| Cross-Artifact Reasoning | 9.4% |
| Combined Framework | 42.6% |

The domain fine-tuning component provides the largest individual contribution (+18.9%), while multi-modal processing (+16.3%) and software knowledge integration (+14.7%) also contribute significantly. The complete framework achieves synergistic improvements (+42.6%) that substantially exceed the sum of individual components.

### E. Comparative Analysis with Existing Tools

Comparison with established software engineering tools demonstrates the advantages of semantic understanding approaches:

TABLE IV
TOOL COMPARISON RESULTS

| Tool Category | Representative Tool | Semantic Mastery Advantage |
|---|---|---|
| Static Analysis | SonarQube | +23% defect detection |
| Code Review | Reviewboard | +31% issue identification |
| Documentation | Doxygen | +38% consistency checking |
| Testing | JUnit Analysis | +27% test coverage analysis |

Semantic Mastery demonstrates consistent advantages over category-specific tools, with particularly strong improvements in documentation consistency checking (+38%) and issue identification during code review (+31%).

## VII. APPLICATION AND USE CASES

### A. Automated Software Engineering Workflows

The Semantic Mastery framework enables several automated workflows that improve software development efficiency and quality:

Intelligent Code Review: Automated analysis of code changes that considers not only syntactic correctness but also semantic consistency with existing codebase, adherence to architectural principles, and potential impact on system behavior. The system provides detailed feedback on code quality, suggests improvements, and identifies potential integration issues.

Requirements Impact Analysis: Automated tracing of requirements changes to affected code components, test cases, and documentation. This capability enables development teams to quickly assess the scope and complexity of feature requests or bug fixes.

Technical Debt Assessment: Comprehensive analysis of codebase quality that identifies areas requiring refactoring, documents architectural violations, and provides prioritized recommendations for improvement activities.

### *B. Development Team Collaboration Enhancement*

The framework provides several capabilities that improve collaboration and knowledge sharing among development teams:

Knowledge Transfer Support: Automated generation of system summaries, component explanations, and developer onboarding materials that help new team members understand complex software systems more quickly.

Cross-Team Interface Analysis: Analysis of API boundaries and integration points between components developed by different teams, identifying potential compatibility issues and suggesting interface improvements.

Documentation Maintenance: Automated detection of outdated documentation, inconsistent specifications, and missing coverage areas that require developer attention.

### *C. Quality Assurance and Testing Support*

Semantic Mastery provides enhanced capabilities for quality assurance activities:

Test Case Generation: Semantic analysis of requirements and code implementations to suggest comprehensive test scenarios, including edge cases and integration testing requirements.

Defect Root Cause Analysis: Analysis of bug reports and failure patterns to identify underlying causes and suggest preventive measures for similar issues.

Regression Testing Optimization: Intelligent selection of test cases based on code changes and their potential impact on system behavior.

## VIII. LIMITATIONS AND FUTURE WORK

### *A. Current Limitations*

Despite demonstrating improved software engineering capabilities, the Semantic Mastery framework faces several limitations that affect its broader applicability:

Scalability Challenges: Analysis of very large codebases (>10M lines of code) requires substantial computational resources and may experience performance degradation. Current implementation focuses on project-level analysis rather than enterprise-scale systems.

Domain-Specific Knowledge Gaps: While the framework performs well on common programming languages and established software engineering practices, it may struggle with highly specialized domains, proprietary technologies, or novel architectural patterns not well-represented in training data.

Dynamic Behavior Understanding: The current framework focuses primarily on static analysis and may not fully capture runtime behavior, performance characteristics, or dynamic system interactions that experienced developers intuitively understand.

### *B. Integration with Development Environments*

Future development will focus on seamless integration with popular development environments and workflows:

IDE Integration: Development of plugins for major integrated development environments (IDEs) that provide real-time semantic analysis and suggestions during code development activities.

CI/CD Pipeline Integration: Enhanced automation capabilities that integrate with continuous integration and deployment pipelines to provide semantic quality gates and automated documentation updates.

Version Control Enhancement: Advanced analysis of code evolution patterns, developer collaboration networks, and change impact assessment integrated with version control systems.

### *C. Advanced Reasoning Capabilities*

Enhancing the framework's reasoning capabilities represents a critical research direction:

Predictive Analysis: Development of capabilities to predict the likely outcomes of proposed code changes, architectural modifications, or design decisions based on historical patterns and system characteristics.

Optimization Recommendations: Advanced analysis that can suggest performance optimizations, architectural improvements, and refactoring strategies based on comprehensive system understanding.

Security Analysis Integration: Incorporation of security-focused analysis capabilities that understand common vulnerability patterns, security best practices, and potential attack vectors in software systems.

## IX. CONCLUSION

The Semantic Mastery framework demonstrates that specialized approaches to enhancing Large Language Models can achieve significant improvements in software engineering applications while providing comprehensive multi-modal analysis capabilities. Our evaluation across multiple software engineering tasks shows substantial performance gains over existing approaches, with particular strengths in defect prediction, requirements traceability, and cross-artifact consistency analysis.

The integration of software engineering-specific knowledge graphs, multi-modal analysis engines, and hierarchical attention mechanisms creates a robust foundation for AI-powered software development support. Real-world deployment results confirm the framework's practical utility in operational development environments, with measurable improvements in code quality, developer productivity, and system maintainability.

The success of semantic understanding approaches in software engineering contexts suggests promising directions for future research in AI-assisted software development. However, the complexity of real-world software systems requires ongoing research into scalability, domain adaptation, and integration with existing development workflows.

Future work should focus on enhancing predictive capabilities, developing more sophisticated reasoning mechanisms, and creating seamless integration with developer toolchains. The ultimate goal is to create AI systems that augment human software engineering expertise rather than replace it, enabling more efficient, reliable, and maintainable software development practices.

The Semantic Mastery framework represents a significant advancement in applying semantic understanding to software engineering challenges, providing a foundation for more intelligent, context-aware, and effective software development tools and processes.